\documentclass[letterpaper,10pt,conference]{ieeeconf}
\IEEEoverridecommandlockouts
\usepackage{amsmath,amssymb,booktabs,graphicx,multirow,url}
\usepackage{newtxtext,newtxmath}
\usepackage{flushend}
\usepackage[hidelinks]{hyperref}
\hypersetup{
  pdftitle={Physical-Touch Observability from Wrist Wrench in Granular Scooping},
  pdfauthor={Hongyi Lin, Song Zhang, Xubo Liu, Yang Liu}
}
\graphicspath{{figures/}}
\newcommand{\ft}{F/T}
\newcommand{\zresp}{z_{\mathrm{resp}}}

\title{\LARGE \bfseries Physical-Touch Observability from Wrist Wrench in Granular Scooping}

\author{%
\authorblockN{Hongyi Lin$^{1}$, Song Zhang$^{2}$, Xubo Liu$^{2}$, and Yang Liu$^{1}$}
\authorblockA{\small
$^{1}$School of Vehicle and Mobility, Tsinghua University\quad
$^{2}$Tsing-AI(Shanghai) Technology Co., Ltd\\
\texttt{hy-lin22@mails.tsinghua.edu.cn}
}
}

\begin{document}
\maketitle

\begin{abstract}
Mining and earthmoving are important real-world deployment settings for embodied intelligence. Autonomous transport and driving systems have improved substantially, but loading and scooping still often depend on skilled human operators, exposing personnel and equipment to operational risk. For robotic scooping, pre-contact RGB-D sensing reveals surface geometry but not the resistance, compaction, tool engagement, or load transfer that emerge during interaction. We test whether the current scoop's six-axis wrist force/torque (F/T), or wrist wrench, contains information about final collected volume, and whether that information depends on the correctly paired action--terrain interaction. We call this property physical-touch observability. Using 6,700 real-robot scoops across 67 terrains, we evaluate correctly paired current-scoop F/T against pre-contact prediction and correspondence-breaking controls under terrain-held-out testing. At the retrospective 60\% sequence boundary, correctly paired F/T reduces mean absolute error by 14.2\% relative to Action-only and by 21.9\% relative to cross-terrain mismatched F/T. Engineered signal summaries reproduce the result across model architectures. Together, these findings position wrist wrench not merely as a low-level feedback signal, but as a task-level perceptual modality through which embodied robots can infer hidden physical states during interaction, providing a foundation for response-aware autonomy in mining and other contact-rich tasks.
\end{abstract}

\section{Introduction}

Loading and scooping are central operations in mining and earthmoving. Reliable autonomy in these tasks depends on understanding how a commanded action interacts with the material and what it is likely to collect. Pre-contact red--green--blue and depth (RGB-D) sensing describes the exposed surface, while the outcome of a scoop also depends on material resistance, packing, tool engagement, and load transfer during contact. These interaction-dependent factors create uncertainty that surface geometry alone cannot resolve. Robotic scooping therefore motivates a broader perception problem: how to obtain task-relevant information from the physical interaction generated by the robot's own action.

Wrist force/torque (F/T), or wrist wrench, provides a mechanically coupled observation of this interaction. Its temporal evolution reflects the combined effects of tool motion, contact forces, and load transfer. Visual and force feedback already support complex manipulation and contact sensing~\cite{monwilliams2025embodied,iskandar2024intrinsic}. For scooping, linking the measured response to final collected volume gives wrist wrench a direct role in task perception. It allows the robot's outcome estimate to incorporate evidence about how the material actually responds to the executed motion. Understanding this information is a foundation for robotic systems that refine their task estimates through physical interaction.

The CoDeGa scooping framework combines pre-contact outcome prediction with adaptation from completed trials to improve subsequent action selection on new terrains~\cite{zhu2023fewshot}. We study the information available within the current scoop. Two questions define the problem. First, does a partial wrist-wrench response contain information that improves final-volume prediction on unseen terrains? Second, does this information depend on the response being correctly paired with the executed action and encountered terrain? The correspondence question is essential because wrist wrench reflects both the commanded motion and the material response. Establishing its task-level value requires distinguishing evidence specific to the current interaction from information that mismatched responses can also supply.

We study \emph{physical-touch observability}: task-outcome information recoverable from the mechanically coupled response of the current scoop. Using 6,700 real-robot scoops across 67 terrains, we evaluate final-volume prediction from partial wrist-wrench recordings under terrain-held-out testing. Pre-contact baselines and correspondence-breaking controls test the value and specificity of the response, while progressively longer observations reveal how outcome information accumulates. The results show that correctly paired wrist responses improve prediction over Action-only, and that mismatched responses fail to reproduce the gain.

Our contributions are:
\begin{itemize}
\item We establish the predictive value of current-scoop wrist wrench for final collected volume under terrain-held-out evaluation.
\item We demonstrate that this value depends on correct action--terrain response correspondence.
\item We characterize information accumulation in partial responses and examine its availability at an observation boundary defined by received sensor samples.
\end{itemize}

\section{Related Work}
\subsection{Granular Scooping and Excavation}
Robot scooping has been studied across granular media, liquids, and food. Learned forward models predict granular-material motion for scoop-and-dump planning~\cite{schenck2017granular}, while curriculum reinforcement learning selects actions for goal-conditioned water scooping~\cite{niu2023goats}. CoDeGa predicts scoop outcomes from pre-contact RGB-D observations and actions, then adapts action selection after completed trials on a new terrain~\cite{zhu2023fewshot}. Food-acquisition systems instead use active interaction or closed-loop visual feedback to stabilize material during scooping~\cite{tai2023scone,grannen2023bimanual}. Excavation controllers also alter motion in response to measured resistance~\cite{franceschini2025oscillatory}, and hydraulic-machine studies estimate external forces and payload from pressure, kinematics, and calibrated dynamics~\cite{kamezaki2012dominant,madau2021online,werner2025calibrated}. These methods improve planning, adaptation, or control; we instead isolate task-outcome information in the mechanical response of the current scoop.

\subsection{Interaction Sensing for Contact-Rich Manipulation}
Contact-generated signals reveal properties that can be ambiguous before contact. Audio-frequency vibration has been used to estimate granular amount and flow~\cite{clarke2018audio}, proprioceptive force to classify excavated materials~\cite{fernando2020material}, and joint torque to localize contacts~\cite{han2023proprioceptive}. In manipulation, action-conditioned models predict tactile evolution and slip~\cite{mandil2022action}, while vision and touch support grasp-outcome prediction, shared representations, and joint inference of kinematic and force trajectories~\cite{calandra2018feeling,lee2019visiontouch,lambert2019joint}. Compact tactile sensors and tactile policies further extend contact-rich perception and control~\cite{lambeta2020digit,oller2023membranes,narang2020interpreting}. Unlike material, contact, or slip estimation, our target is final collected volume, and our controls test whether its prediction requires the correctly paired action--terrain response.

\subsection{Temporal Outcome Modeling and Uncertainty}
Mechanical interaction histories are variable-length multichannel time series. Attention models provide learned sequence representations~\cite{vaswani2017attention}, whereas scalable feature extraction and random convolutional kernels provide structured or computationally inexpensive alternatives~\cite{christ2018tsfresh,dempster2020rocket}. Calibration analysis and deep ensembles provide standard tools for evaluating predictive uncertainty~\cite{guo2017calibration,lakshminarayanan2017ensembles}. We compare a temporal Transformer with engineered observers and audit predictive uncertainty to test the sensing claim across architectures.

\section{Methodology}
\begin{figure*}[!t]
\centering
\includegraphics[width=\textwidth]{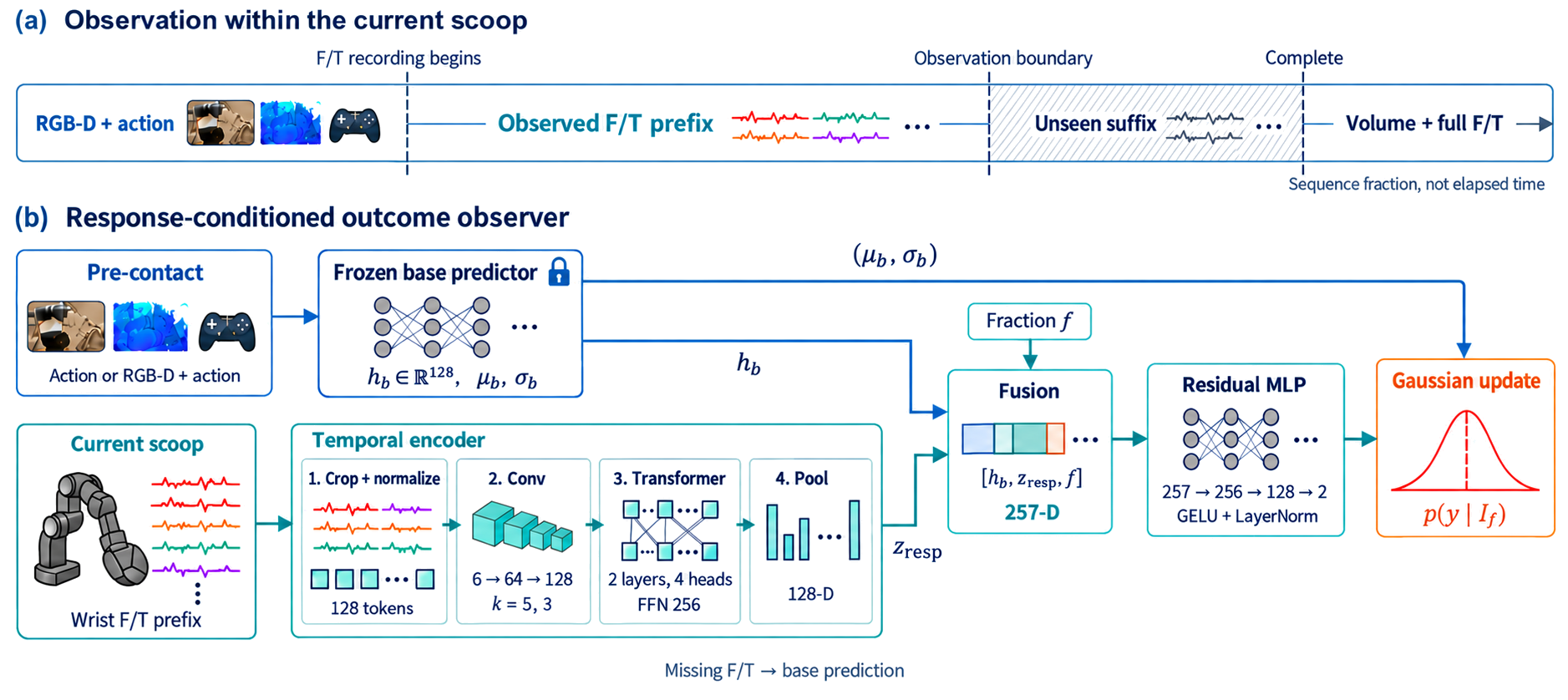}
\caption{Predicting collected volume from a partial wrist F/T sequence. (a) Available observations and unseen suffix. (b) Response-conditioned predictor.}
\label{fig:states}
\end{figure*}

\subsection{Problem Formulation}
Figure~\ref{fig:states} shows the retrospective observation boundary (a) and response-conditioned observer (b). Each scooping trial contains a pre-contact RGB-D observation $o_{\rm RGBD}$, a commanded action $a$, a variable-length six-axis wrist-wrench sequence $w_{1:T}\in\mathbb{R}^{T\times 6}$, and final collected volume $y$. Let $b\in\{A,V\}$ denote the pre-contact base, with $x_A=a$ for the primary Action-only observer and $x_V=(o_{\rm RGBD},a)$ for the visual extension. To study how predictive information accumulates, a fractional observation exposes
\begin{equation}
  \mathcal{I}^{(b)}_f=\{x_b,w_{1:\lceil fT\rceil}\},
  \qquad 0\leq f\leq 1.
  \label{eq:info}
\end{equation}
At $f=0$, the information state contains only the selected pre-contact base input; at $f=1$, it contains the complete response. Intermediate fractions are suffix-free retrospective observations because they depend on the final trace length $T$. Unless a visual input is stated explicitly, reported matched and shuffled F/T contrasts use $b=A$. We use \emph{physical-touch observability} to mean statistical recoverability of task outcome $y$ from the available mechanically coupled response under terrain-held-out evaluation and correct interaction correspondence. It does not imply classical state observability or unique material identification.

\subsection{Response-Conditioned Outcome Observer}
A frozen base network supplies a 128-dimensional latent $h_b$ and Gaussian prediction $(\mu_b,\sigma_b)$. The primary model uses the Action-only base; the visual extension substitutes the frozen RGB-D+Action base. The available wrench prefix is cropped before interpolation, normalized using training-set statistics, resampled to 128 tokens, and encoded by a temporal convolution followed by a two-layer Transformer into the response representation $\zresp\in\mathbb{R}^{128}$. For the retrospective multi-fraction observer, the residual head produces
\begin{align}
 (\delta\mu_f,s_f)&=g([h_b,\zresp,f]),\label{eq:refiner}\\
 \mu_f&=\mu_b+m_{\rm FT}\delta\mu_f,\qquad
 \sigma_f=\sigma_b\exp(m_{\rm FT}s_f),\label{eq:mask}
\end{align}
where $m_{\rm FT}\in\{0,1\}$ is the F/T-availability mask. Missing wrench input therefore yields the unchanged base distribution. The response latent is task-directed: it is not interpreted as a recovered material, hardness, engagement, or hydraulic state.

\subsection{Observation Boundaries and Fallback}
For a stream-observable boundary, we select a sensor-sample count from training-trace lengths:
\begin{equation}
\begin{aligned}
 C(k)&=\frac{\left|\{i\in\mathcal{D}_{\rm train}:v_i=1,\ T_i\geq k\}\right|}{N_{\rm train}},\\
 K_q&=\max\{k:C(k)\geq q\},
\end{aligned}
\label{eq:kq}
\end{equation}
where $v_i$ denotes valid F/T, $N_{\rm train}$ is the number of training trials, and $q$ is the target coverage on training trials. Missing traces remain in the denominator. We use $K_{95}$, $K_{90}$, and $K_{80}$ as shorthand for $K_q$ at $q=0.95$, $0.90$, and $0.80$, respectively. A trial becomes eligible once $K$ sensor samples have arrived. If the trace is missing or ends first, the observer retains the pre-contact prediction:
\begin{equation}
\hat y_i^{(K)}=
\begin{cases}
\hat y_i^{\rm F/T}, & v_i=1\ \text{and}\ T_i\geq K,\\
\hat y_i^{\rm base}, & \text{otherwise}.
\end{cases}
\label{eq:fallback}
\end{equation}
For the prespecified primary boundary $K_{90}$, a separate refiner is trained on $w_{1:K_{90}}$ with no sample-dependent progress input. A retained constant is functionally absorbed into the first-layer bias, giving
\begin{equation}
(\delta\mu_{K_{90}},s_{K_{90}})=g_{K_{90}}([h_b,\zresp^{(K_{90})}]).
\label{eq:fixedrefiner}
\end{equation}
Neither $T$ nor $K/T$ is provided. The $K_{95}$ and $K_{80}$ rows are secondary boundary-sensitivity evaluations obtained by applying the frozen $K_{90}$ refiner to prefixes cropped at the corresponding sample counts; they are not separately trained refiners. The boundary is indexed by received samples rather than seconds because sampling rate and timestamps are unverified. This construction makes the observation available directly from the incoming stream without using final trace length.

\section{Experiments and Results}
\begin{figure*}[!t]
\centering
\includegraphics[width=0.98\textwidth]{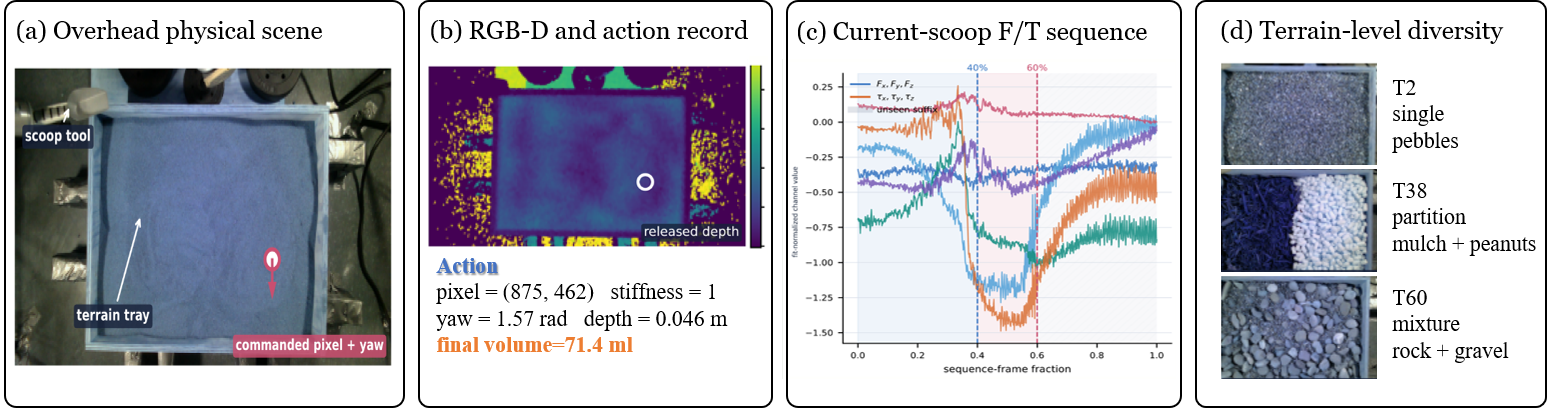}
\caption{Recorded scooping data: (a) RGB image, (b) depth image, action, and final volume, (c) wrist F/T sequence, and (d) terrain examples.}
\label{fig:dataset}
\end{figure*}

\subsection{Experimental Protocol}
\subsubsection{Dataset and split}
We use the public scooping dataset released with CoDeGa~\cite{zhu2023fewshot}, with the terrain-held-out split summarized in Table~\ref{tab:datacontract}. All 6,700 outcomes, including zeros, are retained; 23 trials with missing F/T use an explicit validity mask. Normalization statistics are estimated from the 45 training terrains only, and the five training seeds are 2026--2030. The released action specifies image-plane scoop location, yaw, commanded depth, and a binary stiffness setting. We extract a square RGB-D crop centered on the action location and downsample it from 400 to 100 pixels per side. Figure~\ref{fig:dataset} links an example scoop to its pre-contact images, commanded action, recorded wrench, and final collected volume.

\begin{table}[!ht]
\centering
\caption{Terrain-held-out split and F/T availability.}
\label{tab:datacontract}
\scriptsize
\setlength{\tabcolsep}{3.2pt}
\begin{tabular}{lrrrr}
\toprule
Split & Terrains & Trials & Present & Missing\\
\midrule
Training & 45 & 4,500 & 4,489 & 11\\
Validation & 6 & 600 & 599 & 1\\
OOD test & 16 & 1,600 & 1,589 & 11\\
\midrule
Total & 67 & 6,700 & 6,677 & 23\\
\bottomrule
\end{tabular}
\end{table}

\subsubsection{Baselines and implementation}
Action-only and RGB-D+Action are the pre-contact baselines. Unless labeled RGB-D, the headline F/T comparisons use the Action-only base. The action-conditioned temporal refiner has 391,490 parameters and is trained at $f\in\{0.1,0.2,0.4,0.6,1.0\}$, rotating fractions across epochs so every training trial is observed at every fraction before checkpoint eligibility. The multi-fraction observer minimizes Gaussian negative log-likelihood with Adam and decoupled weight decay (AdamW; learning rate $3\times10^{-4}$, weight decay $10^{-4}$), batch size 64, at most 50 epochs, and validation patience 8. The neural networks were trained using PyTorch on a single NVIDIA GeForce RTX 5080 Laptop GPU. The received-sample analysis uses $K_{95}=657$, $K_{90}=668$, and $K_{80}=691$, all selected from training lengths by Eq.~\eqref{eq:kq}; the $K_{90}$ refiner is trained separately as defined in Eq.~\eqref{eq:fixedrefiner}, whereas $K_{95}$ and $K_{80}$ use the same frozen refiner for boundary-sensitivity evaluation. Training-only normalizers, seeds, terrain split, checkpoint rules, and cache hashes are fixed. Checkpoints are selected on validation terrains before OOD evaluation. Code and evaluation scripts will be publicly released upon acceptance.

For the visual baseline, a four-channel RGB-D encoder uses four stride-2 convolutional blocks (32, 64, 128, and 192 channels), group normalization (GroupNorm), Gaussian error linear unit (GELU) activations, adaptive pooling, and a linear projection to 128 dimensions. A six-dimensional action encoder is fused with that visual latent and mapped to a 128-dimensional frozen base with Gaussian prediction $(\mu_v,\sigma_v)$. RGB-D+Action+F/T applies the same residual formulation to that base. Engineered baselines compute per-channel and force/torque-norm mean, standard deviation, root-mean-square (RMS) value, extrema, range, endpoint change, linear slope, first-difference RMS and maximum, spectral centroid, and low-, mid-, and high-band energy. Ridge regression and histogram-based gradient boosting (HGB) predict the residual from the same frozen base features; the visual engineered variant also receives the visual latent, $\mu_v$, the wrench statistics, and the eligibility indicator. Model and regularization choices use validation terrains only.

The cache stores 128 tokens and six wrench channels. Missing traces are represented by zeros and a validity mask, and raw traces are cropped before interpolation to prevent suffix leakage. We also evaluate force-only and torque-only inputs, as well as reversal, permutation, noise, bias, scale, and dropout perturbations. These perturbations measure the sensitivity of the frozen model.

\subsubsection{Correspondence and temporal controls}
At $K_{90}$, the frozen matched observer receives four F/T conditions: the recipient's own response, another action from the same terrain, the nearest action from another terrain, or a random action from the same paired donor terrain. Recipient action, base features, target, normalization, and checkpoint remain unchanged. Training-only action normalization defines
\begin{align}
 \phi(a)&=[z(x),z(y),z(\sin\theta),z(\cos\theta),z(d),z(s)],\\
 D(a_i,a_j)&=\lVert\phi(a_i)-\phi(a_j)\rVert_2,
\end{align}
where $x,y,\theta,d,s$ denote position, yaw, depth, and stiffness, and $z(\cdot)$ is a z-score using training-action statistics. Twenty no-fixed-point terrain permutations are generated within each split. Nearest-action donors use minimum-cost one-to-one assignment under $D$; random donors use one-to-one random assignment under the same terrain pairing; and same-terrain wrong-action donors use a within-terrain derangement. No donor crosses a split or uses final volume. The mapping-specific eligible cohort is averaged within each seed--terrain cell. An equal-capacity refiner is also trained entirely with nearest-action cross-terrain donor F/T. Twenty fixed training mappings rotate across epochs, each epoch uses 4,065 seeded training draws to match the matched refiner's optimizer-step budget, and checkpoints minimize mean MAE over 20 validation-only mappings. Evaluation uses the 1,313 OOD trials shared by all mappings.

At $f=0.6$, the cross-terrain shuffled-F/T control reuses the same frozen checkpoint and replaces only valid F/T values. Recipient action, base prediction, target, fraction, mask, normalization, and evaluation trial remain unchanged. A deterministic cyclic search selects the first valid same-index donor from another terrain in the same split; missing recipients remain missing, and raw traces are sliced before independent interpolation. Because same index is not established as action-matched, this control can break both terrain and action--response correspondence; the four-condition analysis separates these factors.

Temporal interventions apply deterministic transforms to the available prefix. Mean-only repeats each channel mean. Low-pass retains the lowest 20\% of discrete Fourier-transform bins, high-pass removes them, and eight-block shuffle preserves samples within blocks while disrupting coarse order. Initial and late diagnostics retain 25\% of the available prefix and independently interpolate the segment to 128 tokens. Frequency is expressed in normalized sequence bins rather than physical hertz, and all transformed inputs may introduce distribution shift.

\subsubsection{Metrics and statistical analysis}
We report mean absolute error (MAE) for volume prediction and area under the receiver operating characteristic curve (AUROC) and area under the precision--recall curve (AUPRC) for low-yield ranking. The low-yield threshold is the 25th percentile of positive yields on training terrains, 19.521~mL. The principal reported contrast is 60\% matched F/T versus Action-only; correspondence, visual, model, and temporal comparisons are secondary or diagnostic. For the received-sample and correspondence analyses, MAE is first averaged within terrain; two-sided 95\% confidence intervals (CIs) use 50,000 terrain-bootstrap replicates, with a seed--terrain bootstrap as a sensitivity analysis. The primary 60\% matched-versus-Action-only contrast and the matched-versus-shuffled penalty use 10,000 hierarchical seed--terrain bootstrap replicates: each replicate independently resamples five seeds and 16 terrains from the paired terrain-mean errors. Figures~\ref{fig:curve} and~\ref{fig:interpret} show 95\% Student-$t$ CIs for five-seed means; other fraction and visual contrasts use paired-seed intervals. Both quantify optimization variability on fixed test terrains. Correspondence contrasts are paired within each mapping-specific eligible cohort and averaged across mappings before terrain-level inference. Archived two-sided paired seed-level $t$-tests apply Holm adjustment separately within the visual-control, model-complexity, and temporal-intervention families; their adjusted $p$ values are diagnostic and secondary to the stated intervals. No OOD result selects a boundary, donor rule, model, or hyperparameter.

\subsubsection{Shared-state simulation protocol}
A separate prespecified \emph{arm-minslope-v1} height-field simulation tests paired suffix comparison from an identical restored state. It is uncalibrated, is neither a discrete-element-method simulation nor a robot digital twin, and does not use the real-data observer. A common trajectory reaches 60\% controller progress (Fig.~\ref{fig:synthetic_failure_corners}(a)); the complete integration and random-number states are hashed and restored before seven frozen rules: continue, state-only, force threshold, reactive damping, shuffled-prefix, matched-prefix, and latent oracle. Controller progress differs from the F/T sequence fraction. Five seeds (4101--4105) provide 100 paired episodes per held-out domain: 400 episodes per rule across four prespecified domains. We report composite utility, volume, and force/energy proxies; descriptive intervals resample seeds, domains, and paired episodes.

\subsection{Predictive Information in the Current Response}
\begin{figure}[t]
\centering
\includegraphics[width=\columnwidth]{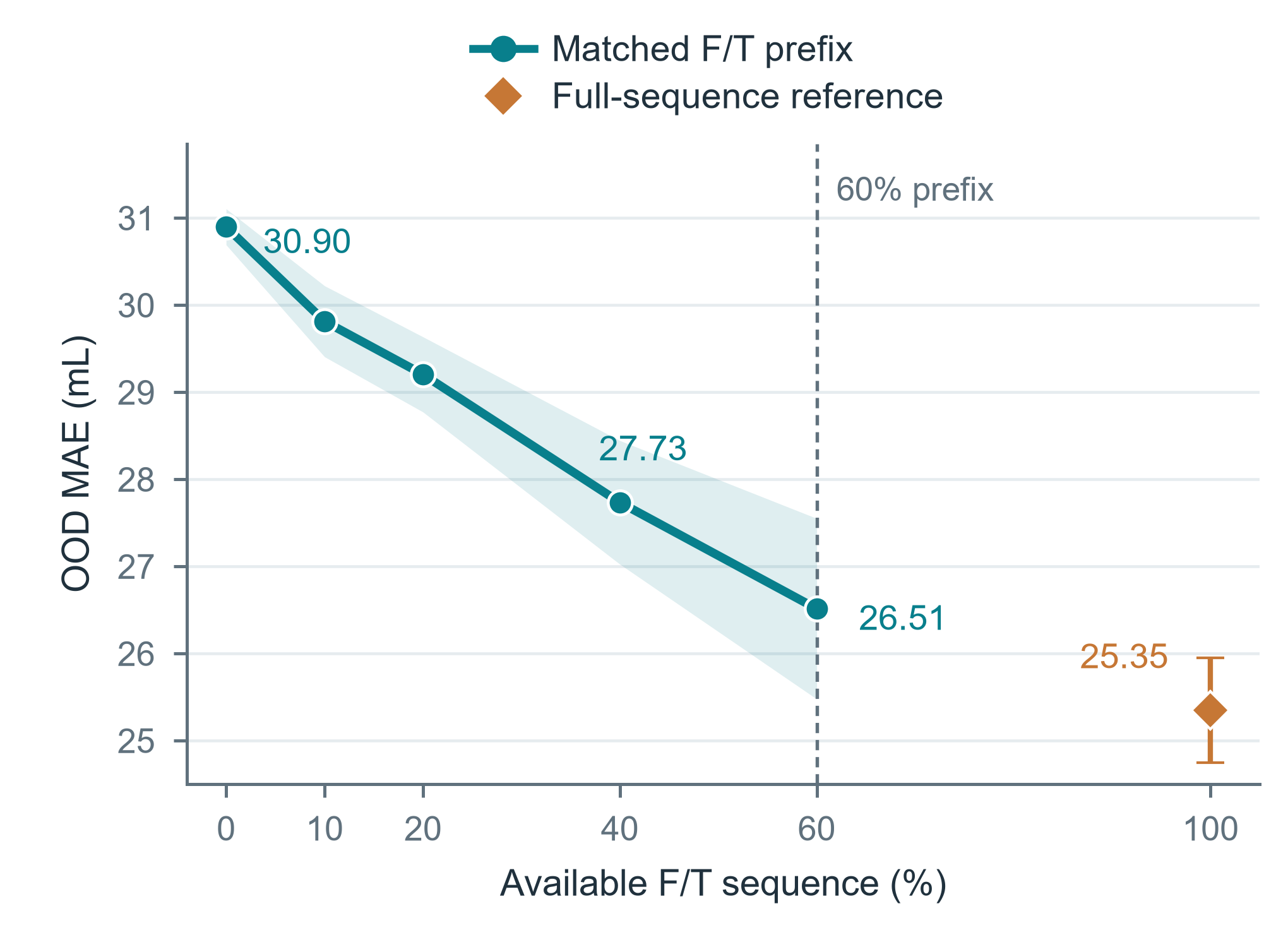}
\caption{Wrist F/T prefixes improve volume prediction on unseen terrains. Shading and the 100\% error bar show 95\% Student-$t$ CIs for five-seed means.}
\label{fig:curve}
\end{figure}

Figure~\ref{fig:curve} and Tables~\ref{tab:main}--\ref{tab:evidencecontract} show how predictive information accumulates: Action+matched F/T reduces MAE from 30.898~mL with no current response to 27.734~mL at 40\% and 26.513~mL at 60\%. The corresponding improvements over Action-only are 3.164~mL at 40\% (paired-seed 95\% CI $[2.544,3.784]$) and 4.385~mL at 60\% (hierarchical seed--terrain 95\% CI $[2.148,6.909]$). The full-sequence reference reaches 25.348~mL, so the 40\% and 60\% observations recover approximately 57\% and 79\% of the Action-only-to-full-sequence MAE reduction.

At 60\%, replacing only the matched F/T response with a cross-terrain donor increases MAE from 26.513 to 33.930~mL, a 7.417~mL penalty (hierarchical seed--terrain 95\% CI $[4.422,11.034]$). Matched response has lower terrain-mean error on 15/16 terrains and lower error in 75/80 terrain--seed cells. On 922 positive-yield OOD trials per seed, matched, Action-only, and shuffled errors are 33.609, 35.921, and 40.666~mL, respectively, showing that zero-volume outcomes do not solely explain the contrast. Because these fractions use final trace length, they characterize retrospective information accumulation.

\begin{table}[t]
\centering
\caption{Terrain-held-out MAE at the reported retrospective prefixes.}
\label{tab:main}
\setlength{\tabcolsep}{3.4pt}
\begin{tabular}{lcc}
\toprule
Method & 40\% (mL) & 60\% (mL)\\
\midrule
Action-only & 30.898 & 30.898\\
RGB-D+Action & 29.255 & 29.255\\
Action+matched \ft{} & 27.734 & 26.513\\
RGB-D+Action+missing \ft{} & 29.255 & 29.255\\
RGB-D+Action+shuffled \ft{} & 29.546 & 29.524\\
RGB-D+Action+matched \ft{} & 28.654 & 27.735\\
\bottomrule
\end{tabular}
\end{table}

Table~\ref{tab:evidencecontract} summarizes the main real-data contrasts. $\Delta$MAE is control MAE minus correctly paired-response MAE, so positive values favor the paired response. Superscripts denote $^{a}$paired-seed intervals, $^{b}$hierarchical seed--terrain bootstrap intervals, and $^{c}$terrain-bootstrap intervals after terrain-first aggregation. The five training seeds share the same 16 held-out terrains and quantify optimization variability.

\begin{table*}[t]
\centering
\caption{Main real-data contrasts.}
\label{tab:evidencecontract}
\scriptsize
\setlength{\tabcolsep}{7pt}
\begin{tabular}{p{10.5cm}cc}
\toprule
Comparison & $\Delta$MAE (mL) & 95\% CI\\
\midrule
Action-only $-$ 60\% matched F/T & 4.385 & $[2.148,6.909]^{b}$\\
Cross-terrain shuffled $-$ 60\% matched F/T & 7.417 & $[4.422,11.034]^{b}$\\
$K_{90}$ same-terrain wrong-action $-$ matched F/T & 6.295 & $[4.285,8.648]^{c}$\\
$K_{90}$ cross-terrain nearest-action $-$ matched F/T & 8.285 & $[6.048,11.093]^{c}$\\
$K_{90}$ cross-terrain random $-$ matched F/T & 8.246 & $[6.023,11.029]^{c}$\\
60\% RGB-D+Action+shuffled F/T $-$ RGB-D+Action+matched F/T & 1.789 & $[1.320,2.258]^{a}$\\
Action-only $-$ $K_{90}$ matched F/T with fallback & 5.036 & $[2.103,8.298]^{c}$\\
\bottomrule
\end{tabular}
\end{table*}

\subsection{Sensitivity to Interaction Correspondence}
Table~\ref{tab:e2} reports five-seed averages over 20 no-fixed-point terrain mappings at $K_{90}$. MAE and $\Delta$MAE are in milliliters, $\Delta$MAE is control minus matched MAE on paired eligible trials, and the intervals are terrain-bootstrap 95\% CIs. Same-terrain wrong-action, cross-terrain nearest-action, and cross-terrain random controls increase MAE relative to the matched response by 6.295, 8.285, and 8.246~mL, respectively. Their terrain-bootstrap 95\% CIs are $[4.285,8.648]$, $[6.048,11.093]$, and $[6.023,11.029]$. Nearest-action donors are substantially closer than random donors in normalized action space (median 1.318 versus 3.389), with 93.75\% versus 50.46\% stiffness agreement and donor reuse of at most one. Cross-terrain donor coverage is 87.6\%, compared with 90.2\% for matched and same-terrain conditions. All 20 mappings retain positive control-minus-matched gaps: their descriptive mean ranges are 5.782--7.037~mL for same-terrain wrong action, 6.519--10.120~mL for cross-terrain nearest action, and 6.326--10.229~mL for cross-terrain random action.

On the strict common OOD cohort of 1,313 trials, an equal-capacity refiner trained only with nearest-action cross-terrain donor F/T shows no resolved improvement over Action-only: Action-only minus donor-trained is $-0.042$~mL (terrain-bootstrap 95\% CI $[-0.170,0.087]$). Donor-trained minus matched-trained is 2.675~mL ($[0.475,4.789]$), so training under broken correspondence does not reproduce the matched-response gain. The latter direction holds on 13/16 terrain means and 64/80 seed--terrain cells. Together, these diagnostics support an interaction-correspondence interpretation.

\begin{table}[t]
\centering
\caption{Correspondence controls at $K_{90}$.}
\label{tab:e2}
\scriptsize
\setlength{\tabcolsep}{1.5pt}
\begin{tabular}{p{0.38\columnwidth}rrr}
\toprule
Condition & MAE (mL) & $\Delta$MAE (mL) & 95\% CI\\
\midrule
Matched & 25.960 & 0 & --\\
Same-terrain wrong-action & 32.255 & 6.295 & [4.285,8.648]\\
Cross-terrain nearest-action & 34.245 & 8.285 & [6.048,11.093]\\
Cross-terrain random & 34.206 & 8.246 & [6.023,11.029]\\
\bottomrule
\end{tabular}
\end{table}

\subsection{Received-Sample Coverage Analysis}
Table~\ref{tab:fixedgate} reports the received-sample analysis, which complements the retrospective series by applying the refiner only after a training-selected number of sensor samples has arrived. It reports OOD coverage and all-trial MAE in milliliters. Base denotes the corresponding frozen pre-contact MAE and Gain is Base minus MAE; 95\% CIs resample terrain means. Gains are computed before the displayed values are rounded, and seeds quantify optimization variability. At $K_{90}=668$, 90.19\% of OOD trials are eligible. For the Action+matched F/T rows, missing or shorter traces retain the unchanged Action-only prediction; the visual-HGB row retains the corresponding frozen RGB-D+Action prediction. Per-terrain coverage ranges from 80\% to 100\%, so no held-out terrain is excluded. Across all 1,600 OOD trials, MAE decreases from 30.898 to 25.862~mL, a 5.036~mL reduction (terrain-bootstrap 95\% CI $[2.103,8.298]$; seed--terrain sensitivity $[2.097,8.323]$), with the same direction on 12/16 terrain means and 61/80 seed--terrain cells. The $K_{95}$ and $K_{80}$ analyses are directionally consistent, with all-trial reductions of 5.350 and 4.575~mL.

Holding the cohort fixed to the 1,272 trials eligible at $K_{80}$ gives terrain-balanced reductions of 6.231, 5.949, and 5.599~mL at $K_{95}$, $K_{90}$, and $K_{80}$, with terrain-bootstrap 95\% CIs of $[2.969,9.718]$, $[2.535,9.639]$, and $[2.148,9.206]$, respectively. The positive reduction at all three boundaries is therefore not explained by changing eligibility.

\begin{table*}[t]
\centering
\caption{Received-sample coverage analysis.}
\label{tab:fixedgate}
\scriptsize
\setlength{\tabcolsep}{4pt}
\begin{tabular}{lrrrrrr}
\toprule
System & Boundary & Coverage & MAE (mL) & Base (mL) & Gain (mL) & 95\% CI \\
\midrule
Action+matched F/T & $K_{95}=657$ & 94.56\% & 25.548 & 30.898 & 5.350 & [2.562, 8.395] \\
Action+matched F/T & $K_{90}=668$ & 90.19\% & 25.862 & 30.898 & 5.036 & [2.103, 8.298] \\
Action+matched F/T & $K_{80}=691$ & 79.50\% & 26.323 & 30.898 & 4.575 & [1.765, 7.566] \\
RGB-D+Action+F/T (HGB) & $K_{90}=668$ & 90.19\% & 27.184 & 29.255 & 2.072 & [0.830, 3.795] \\
\bottomrule
\end{tabular}
\end{table*}

\subsection{Visual Complementarity and Low-Yield Ranking}
RGB-D already explains substantial variation (Table~\ref{tab:main}). At 60\%, RGB-D+Action+matched F/T yields 1.789~mL lower MAE than the identical-capacity shuffled-F/T control (paired-seed 95\% CI $[1.320,2.258]$, Holm-adjusted $p=0.0031$, 5/5 seeds), resolving correspondence-specific information beyond the evaluated pre-contact control. Its 1.520~mL improvement over RGB-D+Action has a paired-seed 95\% CI of $[-0.003,3.042]$ and is not multiplicity-confirmed. Thus the correspondence contrast is resolved, whereas the incremental gain over RGB-D+Action remains uncertain.

Low-yield AUROC/AUPRC rise from 0.643/0.731 at 0\% to 0.722/0.777 at 40\% and 0.739/0.789 at 60\%. At the validation-selected operating point, matched F/T raises recall from 0.555 to 0.778 versus Action-only, but also raises the offline false-positive rate among productive scoops from 0.326 to 0.437. The signal therefore improves ranking, while the preferred operating threshold remains application-cost dependent. Gaussian negative log-likelihood is 4.441 and empirical 90\% interval coverage is 0.835, excluding calibrated-risk claims.

The engineered HGB visual extension combines the frozen RGB-D+Action latent and mean $\mu_v$ with received-prefix F/T statistics and the eligibility indicator. It reduces OOD MAE from 29.255 to 27.184~mL (2.072~mL; terrain-bootstrap 95\% CI $[0.830,3.795]$) and improves 13/16 terrain means. Because it was evaluated after the initial OOD run, this result is exploratory rather than confirmatory.

\begin{figure*}[!t]
\centering
\includegraphics[width=0.98\textwidth]{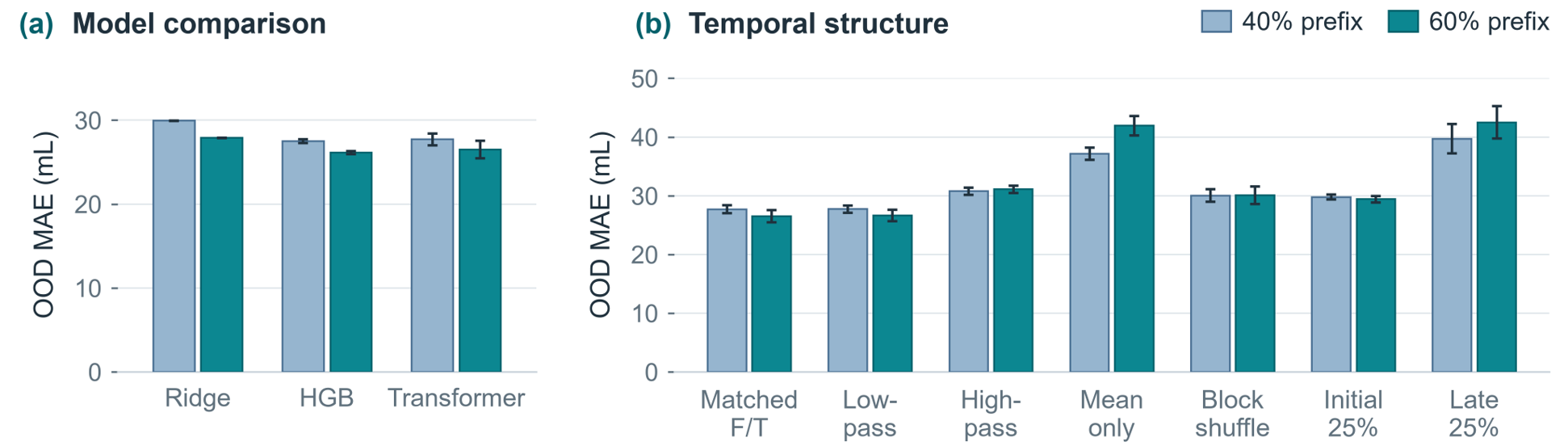}
\caption{OOD prediction error for (a) different models and (b) temporal interventions. Bars compare 40\% and 60\% prefixes; error bars show 95\% Student-$t$ CIs for five-seed means.}
\label{fig:interpret}
\end{figure*}

\begin{figure*}[!t]
\centering
\includegraphics[width=0.98\textwidth]{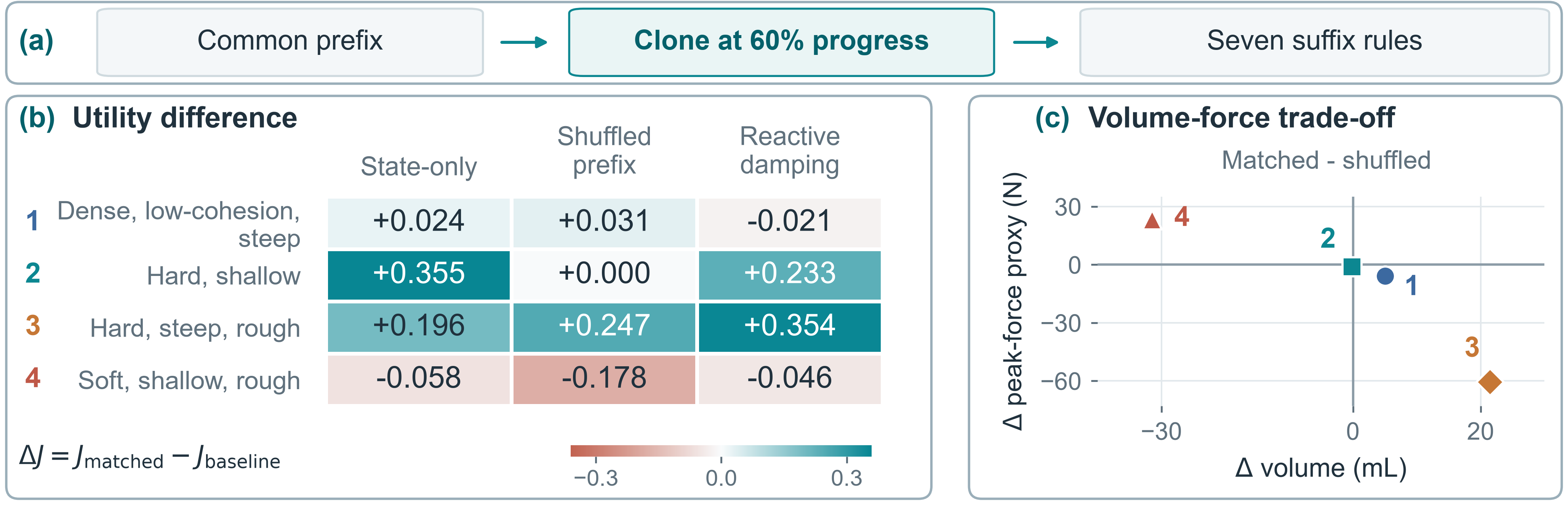}
\caption{Comparing suffix rules from the same simulation state. (a) State restoration and branching. (b) Matched-minus-baseline utility differences. (c) Matched-minus-shuffled volume and peak-force-proxy differences.}
\label{fig:synthetic_failure_corners}
\end{figure*}

\subsection{Model Complexity and Temporal Structure}
Table~\ref{tab:interpretable} and Fig.~\ref{fig:interpret}(a) compare model architectures. Validation selects HGB in all ten seed/fraction runs. At 60\%, engineered HGB obtains 26.154~mL versus 26.513~mL for the Transformer; Transformer-minus-HGB is 0.359~mL with a paired-seed 95\% CI of $[-0.754,1.472]$ and Holm-adjusted $p=0.760$. The comparison does not resolve a performance difference. Linear Ridge is weaker at 27.929~mL. Standardized Ridge coefficients rank slope and endpoint change highest at 60\%, followed by difference RMS, spectral centroid, and mean/RMS. Feature correlation prevents causal attribution, but physically interpretable nonlinear summaries retain most of the predictive information.

\begin{table}[t]
\centering
\caption{OOD MAE for model and temporal comparisons at the 60\% prefix.}
\label{tab:interpretable}
\setlength{\tabcolsep}{3pt}
\begin{tabular}{lc}
\toprule
Input/model & MAE (mL)\\
\midrule
Engineered Ridge & 27.929\\
Engineered HGB & 26.154\\
Temporal Transformer & 26.513\\
\midrule
Transformer, low-pass & 26.660\\
Transformer, high-pass & 31.112\\
Transformer, mean-only & 41.962\\
Transformer, block shuffle & 30.090\\
Initial 25\% segment & 29.437\\
Late 25\% segment & 42.529\\
\bottomrule
\end{tabular}
\end{table}

Low-pass filtering retains nearly all 60\% performance (Fig.~\ref{fig:interpret}(b); MAE 26.660~mL; $\Delta=+0.147$, Holm-adjusted $p=0.093$), whereas high-pass filtering degrades MAE by 4.599~mL (Holm-adjusted $p=0.0058$). Mean-only is worse, and block shuffling adds 3.577~mL error (Holm-adjusted $p=0.0127$). Together with engineered features, these frozen-model diagnostics are consistent with reliance on smooth evolution and coarse ordering rather than a force snapshot or high-frequency component alone. They induce distribution shift and do not identify an optimal physical decomposition. The initial 25\% segment is more useful than the late segment for this frozen model, but resampling and unknown phases preclude a physical-phase claim.

\subsection{Completed-History Replay and Stress Test}
As a cross-task localization diagnostic, completed-history wrench did not improve static-pool action-selection replay (matched-minus-shuffled $+0.027$ attempts, paired-seed 95\% CI $[-0.047,0.101]$). This differs from current-prefix prediction in target, information set, and metric and therefore supplies a complementary rather than matched temporal comparison.

Across four prespecified held-out simulation domains, all matched-policy contrast intervals cross zero. Domain~4 (soft/shallow/rough) reverses direction: matched minus shuffled gives $\Delta J=-0.178$ (Fig.~\ref{fig:synthetic_failure_corners}(b)). Figure~\ref{fig:synthetic_failure_corners}(c) shows the corresponding volume and peak-force-proxy differences. The simulation provides a reproducible paired-branch protocol and a retained failure case that bounds the aggregate result.

\section{Discussion and Limitations}

Current-scoop wrist F/T contains information about final collected volume that generalizes across held-out terrains. Its value depends on the response being correctly paired with the executed action and the encountered terrain. The similar performance of engineered features and the temporal Transformer further suggests that this finding reflects a property of the interaction signal rather than a particular model architecture. In this sense, wrist wrench provides task-relevant evidence that is unavailable from the commanded action or pre-contact geometry alone.

This finding points to a broader role for physical touch in contact-rich robotics. RGB-D sensing describes what the robot can anticipate before contact, whereas wrist wrench reveals what actually happens as the tool engages with the environment. Material resistance, compaction, engagement, and load transfer do not need to be identified separately; their combined effect becomes observable through the mechanically coupled response. We refer to this property as \emph{physical-touch observability}. It treats contact not only as a disturbance to be controlled, but also as an active sensing process that progressively reveals otherwise hidden task state. This perspective may provide a useful bridge between perception and control in granular scooping and other contact-rich tasks where visually similar environments can produce very different physical outcomes.

The experiments establish offline predictive value; closed-loop control improvement remains untested. The evaluation uses one robot dataset and one terrain split; the benefit is broad but not universal across terrains. Missing information about F/T units, sampling rate, synchronization, and contact onset limits physical-frequency, latency, and phase-specific interpretations. The frozen temporal perturbations reveal model reliance but do not establish causal mechanisms, the predictive uncertainty is not yet calibrated for decision making, and the reduced-order simulation does not reproduce particle-scale mechanics or execute the real-data observer. Future work will develop a conditional world model~\cite{lin2026risk} that incorporates physical-touch observations and validate it on a real excavator or loader, examining its ability to support closed-loop decision making in practical earthmoving tasks.

\section{Conclusion}
Correctly paired current-scoop wrist F/T contains terrain-held-out information about final collected volume. Action--terrain substitution shows that generic or mismatched wrench sequences do not reproduce the gain, while engineered features and temporal interventions indicate substantial reliance on smooth evolution and coarse ordering without resolving a difference between HGB and the Transformer. A received-sample analysis with an Action-only fallback further shows that this information remains available at a stream-observable boundary. Together, the results establish physical-touch observability as a measurable property of granular interaction and provide an empirical foundation for response-aware robotic scooping.

\section*{Acknowledgments}
OpenAI's image-generation tool was used to redraw and refine the schematic in Fig.~1 from an author-provided figure.

% Balance the final page after the manuscript content is fixed.
\flushend
\bibliographystyle{IEEEtran}
\bibliography{references}

@inproceedings{zhu2023fewshot,
  author={Yifan Zhu and Pranay Thangeda and Melkior Ornik and Kris Hauser},
  title={Few-shot Adaptation for Manipulating Granular Materials under Domain Shift},
  booktitle={Robotics: Science and Systems},
  year={2023}
}

@inproceedings{schenck2017granular,
  author={Connor Schenck and Jonathan Tompson and Sergey Levine and Dieter Fox},
  title={Learning Robotic Manipulation of Granular Media},
  booktitle={Conference on Robot Learning},
  pages={239--248},
  year={2017}
}

@inproceedings{clarke2018audio,
  author={Samuel Clarke and Travers Rhodes and Christopher G. Atkeson and Oliver Kroemer},
  title={Learning Audio Feedback for Estimating Amount and Flow of Granular Material},
  booktitle={Conference on Robot Learning},
  pages={529--550},
  year={2018}
}

@article{fernando2020material,
  author={Heshan Fernando and Joshua A. Marshall},
  title={What Lies Beneath: Material Classification for Autonomous Excavators Using Proprioceptive Force Sensing and Machine Learning},
  journal={Automation in Construction},
  volume={119},
  pages={103374},
  doi={10.1016/j.autcon.2020.103374},
  year={2020}
}

@inproceedings{mandil2022action,
  author={Willow Mandil and Kiyanoush Nazari and Amir Ghalamzan},
  title={Action Conditioned Tactile Prediction: Case Study on Slip Prediction},
  booktitle={Robotics: Science and Systems},
  doi={10.15607/RSS.2022.XVIII.070},
  year={2022}
}

@inproceedings{niu2023goats,
  author={Yaru Niu and Shiyu Jin and Zeqing Zhang and Jiacheng Zhu and Ding Zhao and Liangjun Zhang},
  title={{GOATS}: Goal Sampling Adaptation for Scooping with Curriculum Reinforcement Learning},
  booktitle={IEEE/RSJ International Conference on Intelligent Robots and Systems},
  pages={1023--1030},
  year={2023}
}

@inproceedings{tai2023scone,
  author={Yen-Ling Tai and Yu Chien Chiu and Yu-Wei Chao and Yi-Ting Chen},
  title={{SCONE}: A Food Scooping Robot Learning Framework with Active Perception},
  booktitle={Conference on Robot Learning},
  pages={849--865},
  year={2023}
}

@inproceedings{grannen2023bimanual,
  author={Jennifer Grannen and Yilin Wu and Suneel Belkhale and Dorsa Sadigh},
  title={Learning Bimanual Scooping Policies for Food Acquisition},
  booktitle={Conference on Robot Learning},
  pages={1510--1519},
  year={2023}
}

@article{calandra2018feeling,
  author={Roberto Calandra and Andrew Owens and Dinesh Jayaraman and Justin Lin and Wenzhen Yuan and Jitendra Malik and Edward H. Adelson and Sergey Levine},
  title={More Than a Feeling: Learning to Grasp and Regrasp Using Vision and Touch},
  journal={IEEE Robotics and Automation Letters},
  volume={3},
  number={4},
  pages={3300--3307},
  doi={10.1109/LRA.2018.2852779},
  year={2018}
}

@article{lee2019visiontouch,
  author={Michelle A. Lee and Yuke Zhu and Peter Zachares and Matthew Tan and Krishnan Srinivasan and Silvio Savarese and Li Fei-Fei and Animesh Garg and Jeannette Bohg},
  title={Making Sense of Vision and Touch: Learning Multimodal Representations for Contact-Rich Tasks},
  journal={IEEE Transactions on Robotics},
  volume={36},
  number={3},
  pages={582--596},
  doi={10.1109/TRO.2019.2959445},
  year={2020}
}

@article{lambeta2020digit,
  author={Mike Lambeta and Po-Wei Chou and Stephen Tian and Brian Yang and Benjamin Maloon and Victoria Rose Most and Dave Stroud and Raymond Santos and Ahmad Byagowi and Gregg Kammerer and Dinesh Jayaraman and Roberto Calandra},
  title={{DIGIT}: A Novel Design for a Low-Cost Compact High-Resolution Tactile Sensor with Application to In-Hand Manipulation},
  journal={IEEE Robotics and Automation Letters},
  volume={5},
  number={3},
  pages={3838--3845},
  year={2020}
}

@inproceedings{oller2023membranes,
  author={Miquel Oller and Mireia Planas i Lisbona and Dmitry Berenson and Nima Fazeli},
  title={Manipulation via Membranes: High-Resolution and Highly Deformable Tactile Sensing and Control},
  booktitle={Conference on Robot Learning},
  pages={1850--1859},
  year={2023}
}

@inproceedings{vaswani2017attention,
  author={Ashish Vaswani and Noam Shazeer and Niki Parmar and Jakob Uszkoreit and Llion Jones and Aidan N. Gomez and Lukasz Kaiser and Illia Polosukhin},
  title={Attention Is All You Need},
  booktitle={Advances in Neural Information Processing Systems},
  year={2017}
}

@article{christ2018tsfresh,
  author={Maximilian Christ and Nils Braun and Julius Neuffer and Andreas W. Kempa-Liehr},
  title={Time Series Feature Extraction on Basis of Scalable Hypothesis Tests ({tsfresh}---A Python Package)},
  journal={Neurocomputing},
  volume={307},
  pages={72--77},
  year={2018}
}

@article{dempster2020rocket,
  author={Angus Dempster and Fran{\c{c}}ois Petitjean and Geoffrey I. Webb},
  title={{ROCKET}: Exceptionally Fast and Accurate Time Series Classification Using Random Convolutional Kernels},
  journal={Data Mining and Knowledge Discovery},
  volume={34},
  pages={1454--1495},
  year={2020}
}

@inproceedings{guo2017calibration,
  author={Chuan Guo and Geoff Pleiss and Yu Sun and Kilian Q. Weinberger},
  title={On Calibration of Modern Neural Networks},
  booktitle={International Conference on Machine Learning},
  pages={1321--1330},
  year={2017}
}

@inproceedings{lakshminarayanan2017ensembles,
  author={Balaji Lakshminarayanan and Alexander Pritzel and Charles Blundell},
  title={Simple and Scalable Predictive Uncertainty Estimation Using Deep Ensembles},
  booktitle={Advances in Neural Information Processing Systems},
  year={2017}
}

@inproceedings{lambert2019joint,
  author={Alexander S. Lambert and Mustafa Mukadam and Balakumar Sundaralingam and Nathan D. Ratliff and Byron Boots and Dieter Fox},
  title={Joint Inference of Kinematic and Force Trajectories with Visuo-Tactile Sensing},
  booktitle={IEEE International Conference on Robotics and Automation},
  pages={3165--3171},
  doi={10.1109/ICRA.2019.8794048},
  year={2019}
}

@article{han2023proprioceptive,
  author={Seo Wook Han and Min Jun Kim},
  title={Proprioceptive Sensor-Based Simultaneous Multi-Contact Point Localization and Force Identification for Robotic Arms},
  journal={arXiv preprint arXiv:2303.03903},
  year={2023}
}

@inproceedings{narang2020interpreting,
  author={Narang, Yashraj S. and Van Wyk, Karl and Mousavian, Arsalan and Fox, Dieter},
  title={Interpreting and Predicting Tactile Signals via a Physics-Based and Data-Driven Framework},
  booktitle={Robotics: Science and Systems},
  doi={10.15607/RSS.2020.XVI.084},
  year={2020}
}

@article{kamezaki2012dominant,
  author={Mitsuhiro Kamezaki and Hiroyasu Iwata and Shigeki Sugano},
  title={Identification of Dominant Error Force Component in Hydraulic Pressure Reading for External Force Detection in Construction Manipulator},
  journal={Journal of Robotics and Mechatronics},
  volume={24}, number={1}, pages={95--104},
  doi={10.20965/jrm.2012.p0095},
  year={2012}
}

@article{madau2021online,
  author={Riccardo Madau and Daniele Colombara and Addison Alexander and Andrea Vacca and Luigi Mazza},
  title={An Online Estimation Algorithm to Predict External Forces Acting on a Front-End Loader},
  journal={Proceedings of the Institution of Mechanical Engineers, Part I: Journal of Systems and Control Engineering},
  volume={235},
  number={9},
  pages={1678--1697},
  doi={10.1177/09596518211005583},
  year={2021}
}

@article{werner2025calibrated,
  author={Lennart Werner and Pol Eyschen and Sean Costello and Pierluigi Micarelli and Marco Hutter},
  title={Calibrated Dynamic Modeling for Force and Payload Estimation in Hydraulic Machinery},
  journal={Construction Robotics},
  volume={9},
  pages={26},
  doi={10.1007/s41693-025-00169-7},
  year={2025}
}

@inproceedings{franceschini2025oscillatory,
  author={Noah Franceschini and Pranay Thangeda and Melkior Ornik and Kris Hauser},
  title={Autonomous Excavation of Challenging Terrain using Oscillatory Primitives and Adaptive Impedance Control},
  booktitle={2025 IEEE International Conference on Robotics and Automation (ICRA)},
  pages={10394--10400},
  doi={10.1109/ICRA55743.2025.11128330},
  year={2025}
}

@inproceedings{lin2026risk,
  author    = {Hongyi Lin and Wenxiu Shi and Heye Huang and Dingyi Zhuang and Song Zhang and Yang Liu and Xiaobo Qu and Jinhua Zhao},
  title     = {Risk-Controllable Multi-View Diffusion for Driving Scenario Generation},
  booktitle = {Proceedings of the IEEE/CVF Conference on Computer Vision and Pattern Recognition (CVPR) Workshops},
  pages     = {5169--5178},
  month     = jun,
  year      = {2026}
}

@article{monwilliams2025embodied,
  author  = {Ruaridh Mon-Williams and Gen Li and Ran Long and Wenqian Du and Christopher G. Lucas},
  title   = {Embodied Large Language Models Enable Robots to Complete Complex Tasks in Unpredictable Environments},
  journal = {Nature Machine Intelligence},
  volume  = {7},
  number  = {4},
  pages   = {592--601},
  year    = {2025},
}

@article{iskandar2024intrinsic,
  author  = {Maged Iskandar and Alin Albu-Sch{\"a}ffer and Alexander Dietrich},
  title   = {Intrinsic Sense of Touch for Intuitive Physical Human--Robot Interaction},
  journal = {Science Robotics},
  volume  = {9},
  number  = {93},
  pages   = {eadn4008},
  month   = aug,
  year    = {2024},
}
\end{document}